\documentclass[letterpaper, 10 pt, conference]{ieeeconf}  

\IEEEoverridecommandlockouts                              
\usepackage{dcolumn}
\usepackage{bm}
\usepackage{amsmath}
\usepackage{ifthen}
\usepackage{amssymb}
\usepackage[pdftex]{graphicx}
\usepackage{color}

\title{\LARGE \bf
Multiple Kernel Learning for Brain-Computer Interfacing
}

\author{Wojciech Samek$^{1}$, Alexander Binder$^{1}$, Klaus-Robert M\"uller$^{1,2}$
\thanks{$^{1}$W. Samek ({\tt\small wojciech.samek@tu-berlin.de}),\newline
A. Binder ({\tt\small alexander.binder@tu-berlin.de}),\newline
K.-R. M\"uller ({\tt\small klaus-robert.mueller@tu-berlin.de}),
are with the Berlin Institute of Technology, Marchstr.~23, 10587 Berlin, Germany.}%
\thanks{$^{2}$K.-R. M\"uller is with the Department of Brain and Cognitive Engineering, Korea University, Anam-dong, Seongbuk-gu, Seoul 136-713, Korea}%
\thanks{*This work was supported by the German Research Foundation (GRK 1589/1), by the Federal Ministry of Education and Research (BMBF) under the project Adaptive BCI (FKZ 01GQ1115) and by the World Class University Program through the National Research Foundation of Korea funded by the Ministry of Education, Science, and Technology, under Grant R31-10008.}
}

\usepackage{times}
\usepackage{epsfig}
\usepackage{graphicx}
\usepackage{amsmath}
\usepackage{amssymb}
\usepackage{xcolor}
\usepackage{subfig}
\usepackage{url}
\usepackage[normalem]{ulem}
\usepackage{wrapfig}

\newcommand{\vbeta}{{\boldsymbol \beta}}
\newcommand{\vx}{{\boldsymbol x}}
\newcommand{\vw}{{\boldsymbol w}}

\newcommand{\valpha}{{\boldsymbol \alpha}}

\renewcommand{\k}{k}

\newcommand{\ind}{l}
 
\newcommand{\betaj}{\beta}
\newcommand{\matW}{{\boldsymbol W}}
\newcommand{\matC}{{\boldsymbol C}}
\newcommand{\matX}{{\boldsymbol X}}

\renewcommand{\vec}[1]{\ensuremath{\boldsymbol{#1}}}

\begin{document}
\maketitle
\thispagestyle{empty}
\pagestyle{empty}

\begin{abstract}
Combining information from different sources is a common way to improve classification accuracy in Brain-Computer Interfacing (BCI).
For instance, in small sample settings it is useful to integrate data from other subjects or sessions in order to improve the estimation quality of the spatial filters or the classifier. Since data from different subjects may show large variability, it is crucial to weight the contributions according to importance. Many multi-subject learning algorithms determine the optimal weighting in a separate step by using heuristics, however, without ensuring that the selected weights are optimal with respect to classification.
In this work we apply Multiple Kernel Learning (MKL) to this problem. MKL has been widely used for feature fusion in computer vision and allows to simultaneously learn the classifier and the optimal weighting. We compare the MKL method to two baseline approaches and investigate the reasons for performance improvement.
\end{abstract}

\section{Introduction}
Extracting robust and informative features from data is a crucial step for successful decoding of the user's intention in Brain-Computer Interfacing (BCI) \cite{DorMilHinFarMue07}. One of the most popular feature extraction methods in BCI is Common Spatial Patterns (CSP) \cite{Blankertz08optimizingspatial}. It is well suited to discriminate between different mental states induced by motor imagery as it enhances the ERD/ERS effect \cite{DorMilHinFarMue07} by maximizing the differences in band power between two conditions.
Since CSP is a data driven approach it is prone to overfitting and may provide suboptimal results if data is scarce, non-stationary or affected by artefacts. Recently, several extensions have been proposed to robustify the algorithm, e.g.\ \cite{LotGua11, SamJNE12, kan09, TomMue09, Lotte2010, Dev11, SamARX12}. One of the strategies to improve the estimation quality of the spatial filters is to utilize data from other subjects, e.g.\ by regularizing the estimated covariance matrix towards the covariance matrices of other users. However, it has been shown \cite{SamARX12} that inclusion of other subjects' data may harm CSP performance if the discriminative subspaces of the different data sets do not match. Therefore it is crucial to weight the contributions from other users according to importance\footnote{Note that this also includes the exclusion of some subjects.}. The optimal weighting is usually computed in a separate step by applying a heuristic, e.g.\ the composite CSP method (cCSP) \cite{kan09} uses weights that are proportional to the similarity, measured by Kullback-Leibler divergence, between subjects. Note that such heuristics do not ensure that the selected weights are optimal with respect to classification.

In this work we apply Multiple Kernel Learning (MKL) \cite{Lanckriet04, Klo11} to the data integration problem in BCI.
This method allows to simultaneously learn the classifier and the optimal weighting and has been successfully applied to the problem of feature fusion in computer vision \cite{BinPLOS12}. Note that the MKL approach has been applied to BCI before \cite{Hua11}, but in a completely different scenario, namely as single subject classifier with different kernels and not for solving the data integration problem.

This paper is organized as follows. In the next section we present the Multiple Kernel Learning method and its application to BCI.
In Section III we compare it with two baseline approaches on a data set of 30 subjects performing motor imagery. 
We conclude in Section IV with a discussion.

\section{Multiple Kernel Learning for BCI}
\subsection{Multiple Kernel Learning}
Support Vector Machines (SVM) \cite{Burges1998, MueMikRaeTsuSch01} are binary classifiers that learn a linear decision hyperplane with a separating margin between the two classes e.g.\ left and right hand motor imagery. They have been widely used in many different areas and can be extended to non-linear decision boundaries by applying the ``kernel trick''. The SVM decision function can be written as 
\begin{equation}
f(\vx^{new}) = \sum_{i=1}^n \alpha_i k(\vx^i,\vx^{new}) + b,
\label{eq:mkl}
\end{equation}
where $\vx^{new}$ is the trial to be classified, $\vx^i$ is a training trial, $k(\cdot,\cdot)$ denotes the kernel function and $\alpha_i$ and $b$ are parameters which are determined when training the SVM. The integration of data from different sources can be achieved by computing a kernel for each source and combining them.
The decision function in this case is 
\begin{equation}
f(\vx^{new}) = \sum_{i=1}^n \alpha_i \sum_{j=1}^m \beta_jk_j(\vx^i,\vx^{new}) + b,
\label{eq:mkl2}
\end{equation}
where $\beta_j \geq 0$ are the kernel weights assigning an importance value to each source $j$.
Multiple Kernel Learning (MKL) \cite{Lanckriet04} simultaneously optimizes for the parameters $\alpha_i$, $b$ and $\beta_j$. Note that the degree of sparsity of the weight vector $\vec \beta = [\beta_1 \ldots \beta_m]$ can be controlled by adding a $\ell_p$-norm constraint $|| \vec \beta ||^p = 1$ (see \cite{Klo11, KloBreSonZieLasMue09} for details).

\subsection{Application to BCI}
The data integration problem in Brain-Computer Interfacing can be solved on different levels.
The simplest approach is to pool data extracted from different sources and to apply the learning algorithms to the combined set. An alternative is to train a model on each data set separately, to apply all models to the data of interest and to combine the classifier outputs (see e.g.\  \cite{DorBlaCurMul04, fazli2009subject, Fazli11}). Finally one can combine the information from different sources on a medium level of representation, namely on the feature level. In this work we propose to perform data integration on this level by computing a set of features and a kernel for each source. Multiple Kernel Learning (MKL) then combines the kernels in a way that is optimal for classification.

\begin{figure*}
\centering
\includegraphics[width=15cm]{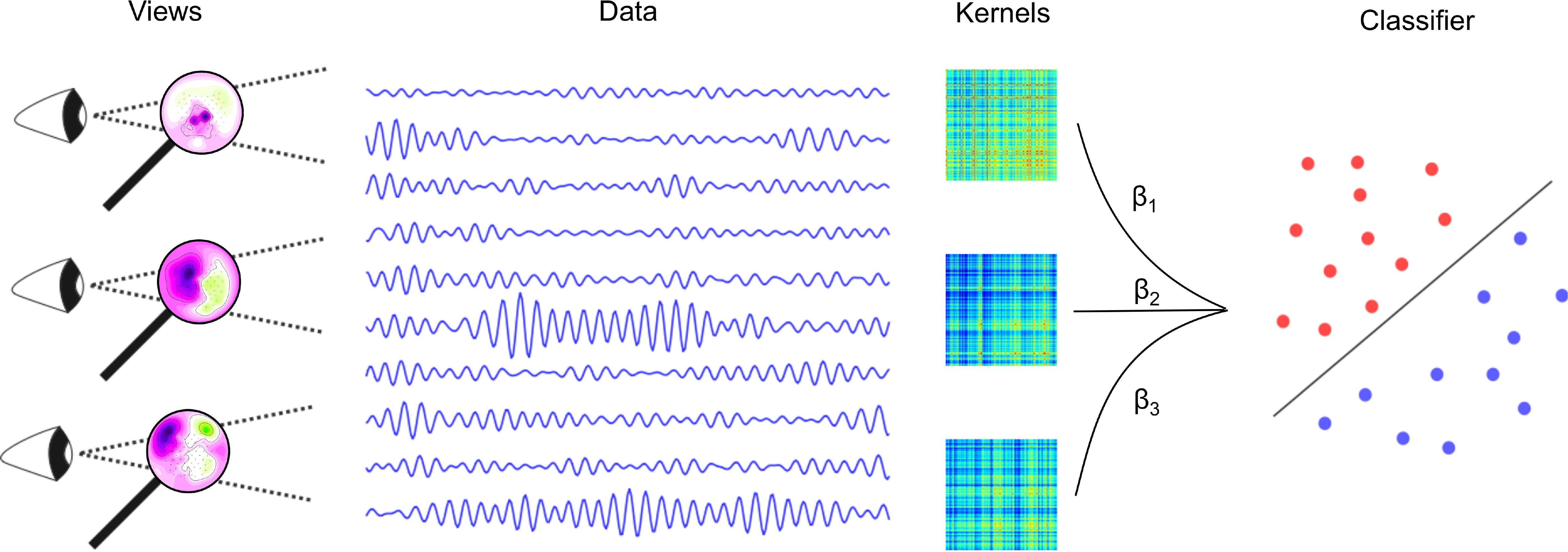}
\caption{Application of Multiple Kernel Learning to BCI. The data integration task consists of combining different views on the data of interest. Looking at the data through ``the lens of'' spatial filters extracted from other users provides a much richer picture of the data. The different sources of information are integrated by computing a kernel for each view and performing a weighted combination of them. The Multiple Kernel Learning algorithm allows to simultaneously train the classifier and optimize the kernel weights $\beta_j$.}
\label{fig:grafik}
\end{figure*}

The application of our method to BCI is summarized by Figure \ref{fig:grafik}. The core idea of it is to provide different ``views'' on the data of interest and to automatically select the important information by combining them in a way that is optimal with respect to classification. 
In the following we describe our method when training a classifier for subject $k$.

In the first step we compute a set of spatial filters $\matW^j = [\vw^j_1, \ldots, \vw^j_6]$ for each subject $j$ (including $k$) by solving the generalized eigenvalue problem
\begin{equation}
\matC_1^j\vw^j_i = \lambda_i\matC_2^j\vw^j_i,
\label{eq:csp}
\end{equation}
and selecting three filters $\vw_i^j$ with largest $\lambda_i$ and three with smallest $\lambda_i$.
Note that $\matC_c^j$ denotes the estimated covariance matrix of class $c$ and subject $j$.
Then we apply these filters (including $\matW^k$) to the data of subject $k$ and compute log-variance features $\vec f_i^j$ 
for each trial $i$ as
\begin{equation}
\vec f_i^j = \log(\rm{var}((\matW^j)^\top\matX^k_i)).
\label{eq:csp}
\end{equation}
Note that $\matX_i^k$ is the band-pass filtered EEG data of trial $i$ and subject $k$.
By using filters from other subjects we look at the data of user $k$ through ``the lens of'' other subjects. This is advantageous when e.g.\ spatial filters can not be reliably computed from subjects' $k$ data because of artefacts or a small-sample setting. Pooling data from all subjects is suboptimal as the different data sets may vary strongly, i.e.\ only the information contained in a subset of other subjects may be relevant. After this feature extraction step we compute a linear kernel matrix for each view $j$ as
\begin{align}
& k_j(\vec f_i^{j},\vec f_l^{j}) = (\vec f_i^{j})^\top \vec f_l^{j}.
\end{align}
The kernels are then combined and a classifier is learned by solving the following MKL optimization problem
\begin{align}
  \min_{\vbeta} \, \max_{\valpha} \quad & \sum_{i=1}^n\alpha_i -\frac{1}{2}\sum_{i,\ind=1}^n\alpha_i\alpha_\ind y_iy_\ind\sum_{j=1}^m\betaj_j\k_j(\vec f_i^{j},\vec f_l^{j}) \\
  \text{s.t.} \quad & \forall_{i=1}^n:\,\, 0\leq\alpha_i\leq C; \quad \sum_{i=1}^n y_i\alpha_i=0; \nonumber \\
&\forall_{j=1}^m: \,\betaj_j\geq 0; \quad \Vert\vbeta\Vert_p \leq 1 \nonumber.
\end{align}
where $m$ is the number of views, $n$ is the number of training trials, $\beta_j$ denotes the kernel weight, $y_i$ and $y_l$ represent trial labels and $\alpha_i, C$ are SVM parameters.
We denote this approach as mklCSP.

\section{Experimental Evaluation}

\subsection{Dataset and Experimental Setup}
The experimental evaluation in this work is based on the Vital BCI data set \cite{Blankertz10} containing EEG recordings from 80 healthy users performing motor imagery with the left and right hand or with the feet. We restrict our analysis to the 30 subjects performing left hand vs.\ right hand motor imagery. The data set contains measurements from a calibration and a test session recorded on the same day, the former consists of 150 trials without feedback and latter consists of 300 trials with 1D visual feedback. All subjects in this study are BCI novices.
The EEG signal was recorded from 119 Ag/AgCl electrodes, band-pass filtered between 0.05 and 200 Hz and downsampled to 100 Hz. We manually select a set of 62 electrodes densely covering the motor cortex and extract a time segment located from 750ms to 3500ms after the cue instructing the subject to perform motor imagery. Furthermore we band-pass filter the data in 8-30 Hz using a 5-th order Butterworth filter and use six spatial filters.

We compare three algorithms in the experiments, namely CSP, cCSP \cite{kan09} and our novel mklCSP approach. 
Note that CSP does not perform data integration and the composite CSP (cCSP) method incorporates data from other subjects by regularizing the covariance matrix $\matC^k_c$ as $$\tilde{\matC}_c^k = (1-\lambda)\matC_c^k + \lambda\sum_{j=1, j\not= k}^m\alpha_{j}  \matC_c^j$$ with $\alpha_{j} = \frac{1}{Z}\cdot \frac{1}{KL[\matC_c^j || \matC_c^k]},\ $ $Z = \sum_{l \not= k} \frac{1}{KL[\matC_c^l || \matC_c^k]}$ and $KL[\cdot || \cdot]$ is the Kullback-Leibler Divergence\footnote{The Kullback-Leibler Divergence between Gaussians is defined as $D_\text{KL}(\mathcal{N}_0 \| \mathcal{N}_1) = { 1 \over 2 } \left( \mathrm{tr} \left( \Sigma_1^{-1} \Sigma_0 \right) + \left( \mu_1 - \mu_0\right)^\top \Sigma_1^{-1} ( \mu_1 - \mu_0 ) -\ln \left( { \det \Sigma_0 \over \det \Sigma_1  } \right) - k  \right).$}.
In order to allow better comparison we apply two types of classifiers after filtering the data with CSP and cCSP, namely Linear Discriminant Analysis (LDA) and Support Vector Machine (SVM).
We use 5-fold cross-validation on the training data to select the relevant parameters and apply lowest error rate as selection criterion. Our algorithm has two free parameters, namely the SVM regularization parameter $C$ and the norm parameter $p$. We select $C$ from $10^{i}$ with $i \in \{-2, -1.5, \ldots, 1.5, 2\}$ and $p$ from $\{1, 1.125, 1.333, 2, \infty \}$ (as done in \cite{BinPLOS12}). We normalize the kernels by the average diagonal value. For the cCSP method we select $\lambda$ from $\{ 0, 10^{-5}, 10^{-4}, 10^{-3}, 10^{-2}, 0.1, 0.2,\ldots, 1\}$.

\subsection{Results and Evaluation}
\begin{figure}
\centering
\includegraphics[width=7.5cm]{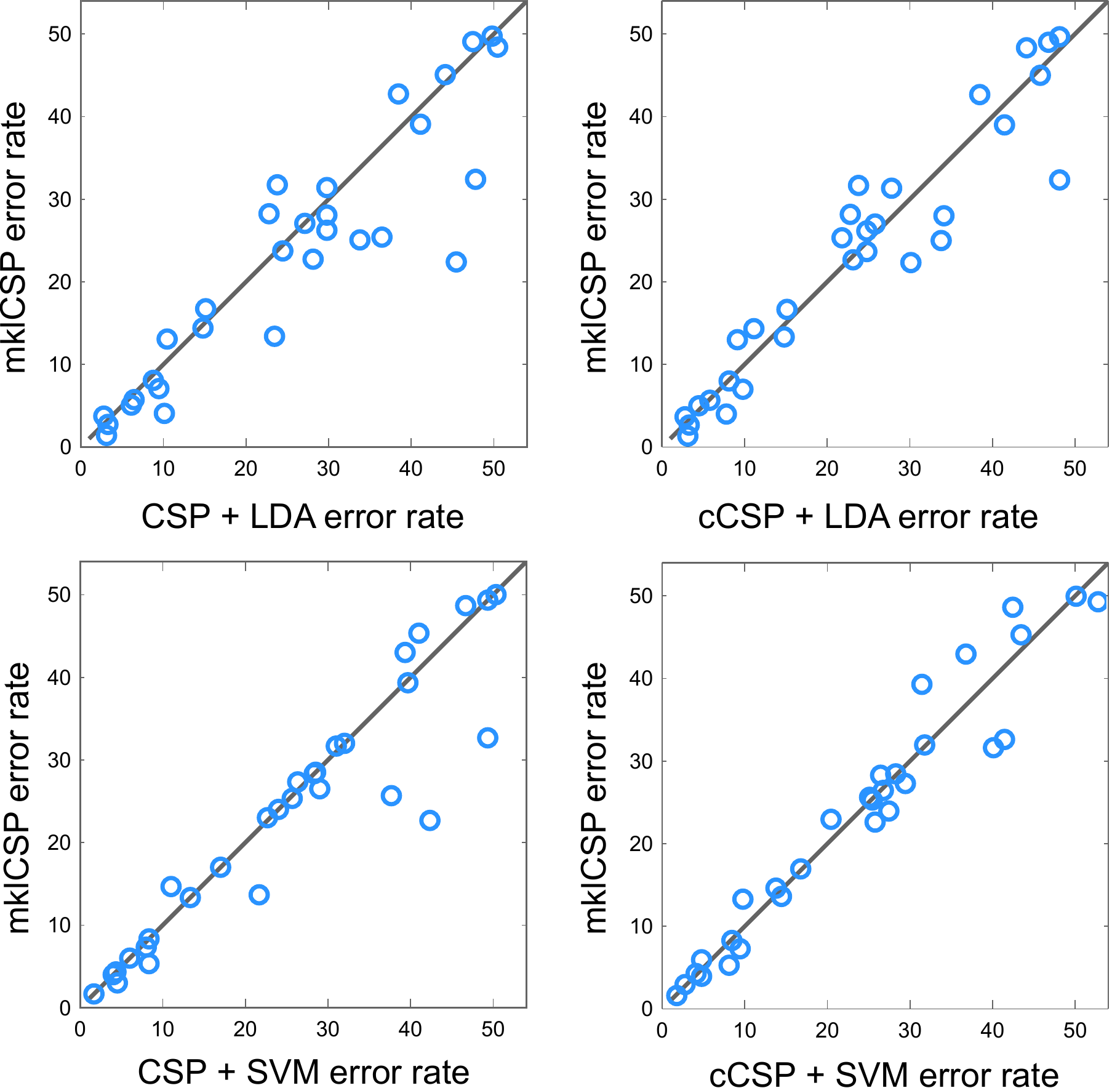}
\caption{Comparison of the error rates of our mklCSP method and the CSP and covCSP baselines using an LDA classifier or SVM. Each circle represents the test error of a subject and if the circle is below the solid line then our method is superior to the baseline.}
\label{fig:res}
\end{figure}

Fig.\ \ref{fig:res} compares the different approaches by using scatter plots. The test error of each subject is represented by a circle. The error rate of the baseline method is represented by the x-coordinate, whereas the y-coordinate denotes the error rate of mklCSP. Thus if the circle is below the solid line than our method performs better than the baseline.
The mklCSP approach is superior to the CSP baseline methods and it is on part to the state-of-the-art cCSP approach that also uses data from other subjects. A potential advantage of our method is that it selects the importance of each subject by optimizing a criterion that is relevant for classification, whereas cCSP uses a similarity heuristic. Furthermore we can control the sparsity of the solution by changing $p$.

\begin{figure}
\centering
\includegraphics[width=8.5cm]{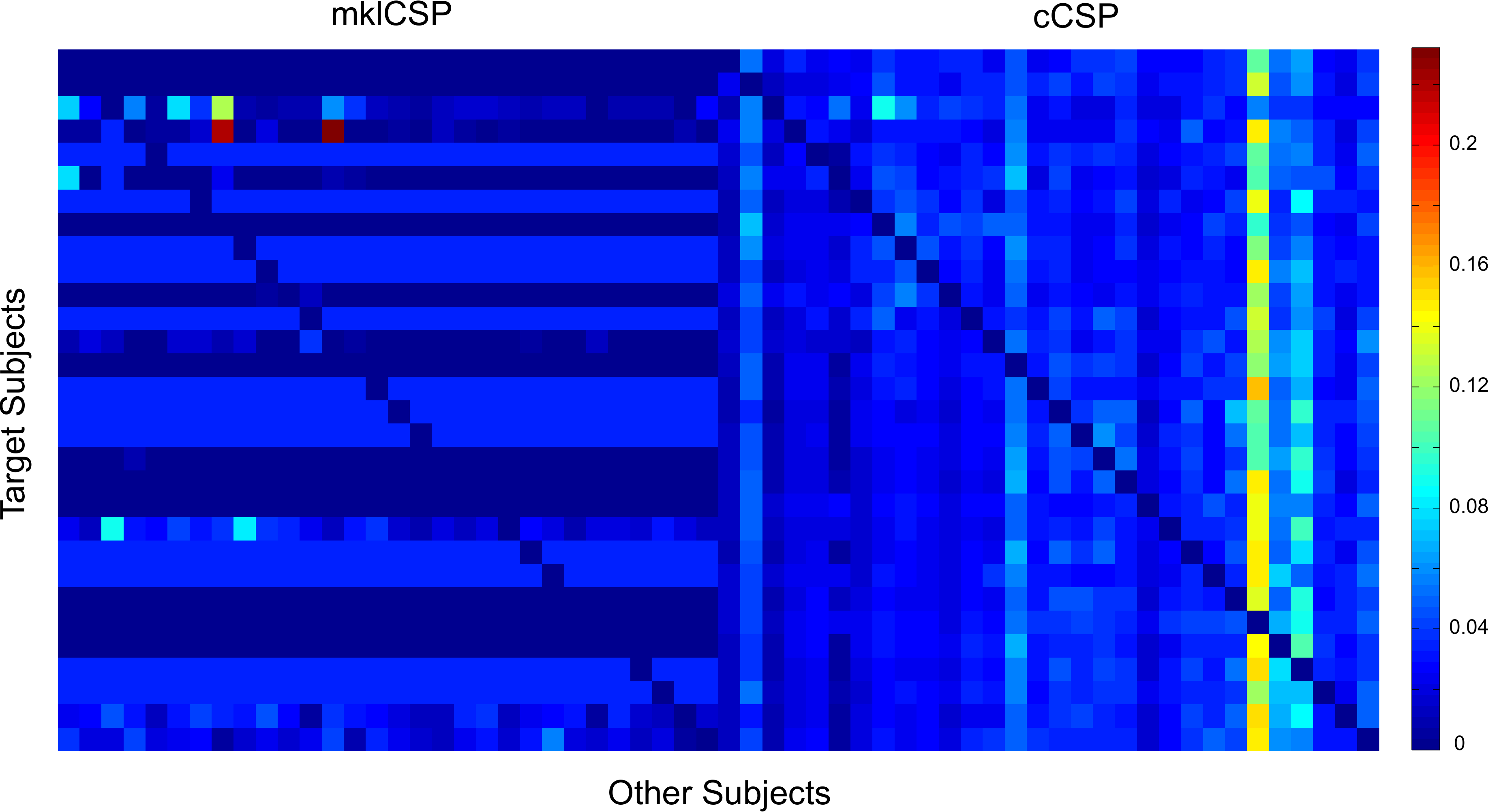}
\caption{MKL kernel weight $\beta_j$ and similarity score $\alpha_j$ matrix. The rows represents the users of interest and the columns stand for the other sources of information. The weights selected by MKL do not correlate with the similarity scores, but are optimized with respect to the objective function of the classification problem.}
\label{fig:kw}
\end{figure}

In the following we investigate the reasons for the improved performance of our method, i.e.\ the advantages of looking at the data through the lens of other subjects. Spatial filters computed on other subjects performing the same motor imagery task may better capture the important information than filters extracted from the data of interest, especially if the data contains artefacts or is very scarce. 
In order to understand what information is transferred between subjects we analyse the MKL kernel weights and the similarity scores $\alpha_j$ of all subjects in Fig.\ \ref{fig:kw}.
The target subjects, i.e.\ the subjects whose motion intentions are decoded and classified, are shown on the rows, whereas users representing the additional sources of information are shown on the columns. We set the diagonal elements of the two matrices to zero for better visibility. In the case of mklCSP one can see that many users, e.g.\ the second one, prefer sparse MKL solutions and do not incorporate information from other subjects. On the other hand there are users that show some strong contributions from one or two subjects and yet others apply relatively small weights to all sources of information. Note that the MKL weights do not correlate with the similarity values $\alpha_j$ which are shown in the right matrix. There is one subject (user 25) that seems to be very similar to other users, i.e.\ the divergence between his covariance matrix and the covariance matrices of other users is small. This means that his impact on the other participants is relatively large, however, note that the right matrix in Fig.\ \ref{fig:kw} only shows the similarity and not the final weights $\lambda\alpha_j$.

\begin{figure}
\centering
\includegraphics[width=8.5cm]{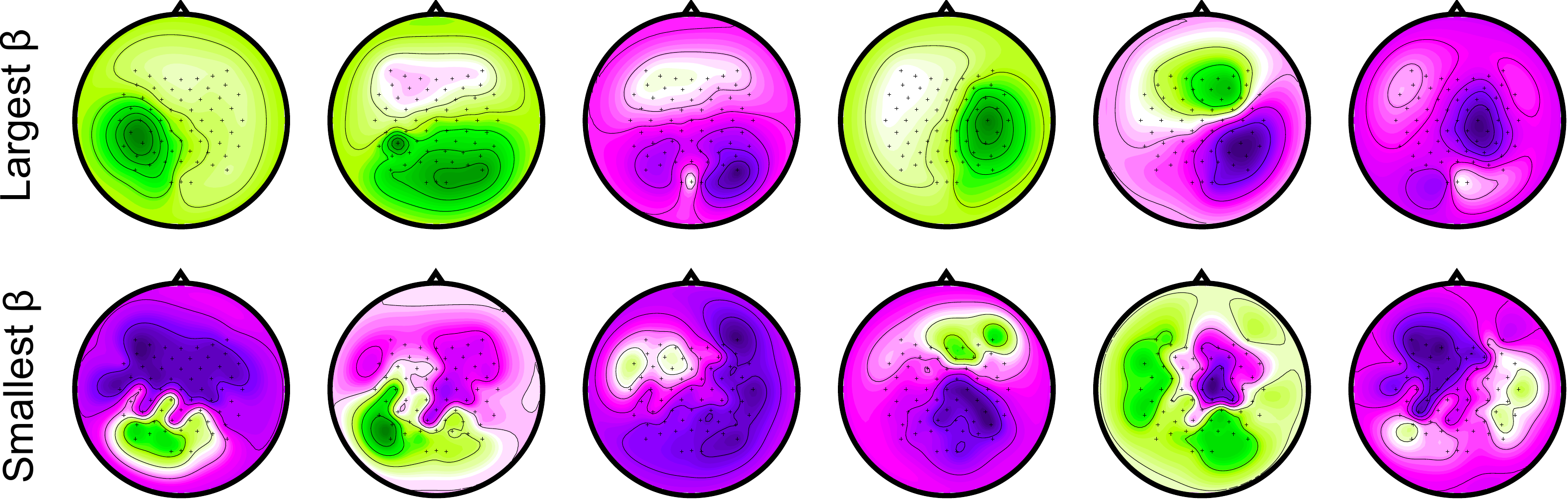}
\caption{Upper row: Activity patterns of the subject that received the largest attention by mklCSP, i.e.\ the largest average weight $\beta_j$. The first and fourth pattern in the upper row shows clear motor imagery activity. Lower row: Activity patterns that received the lowest attention, i.e.\ the smallest average kernel weight. This set of patterns does not show the relevant activity.}
\label{fig:pat}
\end{figure}

In order to investigate what makes a good and bad set of spatial filters we average the MKL weights over all subjects and 
visualize the activity patterns of the most attractive filters, i.e.\ the view with the largest average $\beta_j$, and the least attractive ones, i.e.\ the patterns that correspond to the kernel with smallest average $\beta_j$.
The upper row of Fig.\ \ref{fig:pat} shows the activity patterns of the subject with the largest weight. One can clearly see that the first and fourth patterns show activity which is related to right and left hand motor imagery, thus it makes sense to apply the corresponding spatial filters to the data of interest. On the other hand the lower row of Fig.\ \ref{fig:pat} shows the patterns which were not so useful for the other subjects. Note that these patterns are not very clean, thus they do not perform very well and were not selected by mklCSP.

\section{Discussion}
We showed that Multiple Kernel Learning can be applied to the data integration problem in BCI.
It does not rely on heuristics, but rather automatically determines the importance of other data sets in a way that is optimal for classification.
Using spatial filters computed on other subjects may significantly improve classification accuracy, however, weighting the data according to importance is crucial when using multi-subject approaches.

In future research we would like to apply MKL to other data fusion problems in BCI. For instance, the proposed method can be used to find the best combination of narrow frequency bands for a particular subject. In this case one would look at the data not from the lens of other users, but from the perspective of different frequency bands. First experiments show promising results. Furthermore we plan to investigate the impact of the kernel on classification. As in computer vision we expect that Brain-Computer Interfacing may profit from using non-linear classifiers.

\bibliographystyle{IEEEtran}
\bibliography{bme2012}

\end{document}